%% file: main.tex
\documentclass[runningheads]{styles/llncs}

\usepackage[mobile]{styles/eccv}

\usepackage{styles/eccvabbrv}

\usepackage{graphicx}
\usepackage{booktabs}
\usepackage{amsmath}
\usepackage{mathtools}
\usepackage{empheq}
\usepackage{colortbl}

\usepackage[accsupp]{axessibility}  

\usepackage{hyperref}

\hypersetup{%
  colorlinks=true,
  urlbordercolor={0 1 0},
  citebordercolor={green},
  linkcolor={magenta},
  filecolor=magenta,
  filebordercolor={green},      
  urlcolor=magenta,
  raiselinks=false
}

\usepackage{orcidlink}

\input{src/preamble}
\input{includes/custom_commands}

\begin{document}

\title{\name: Gradient Origin Embeddings for Representation Agnostic 3D Feature Learning
}

\titlerunning{\name}

\author{
Animesh Karnewar\inst{1,2} \and 
Roman Shapovalov\inst{2} \and 
Tom Monnier\inst{2} \and 
Andrea Vedaldi\inst{2} \and 
Niloy J. Mitra\inst{1} \and 
David Novotny\inst{2} 
}

\authorrunning{A.~Karnewar et al.}

\institute{University College London, United Kingdom \and
MetaAI London, United Kingdom\\
}

\maketitle
\thispagestyle{empty}
\begin{center}
\centering
\vspace{-0.5cm}
\url{https://holodiffusion.github.io/goembed}
\vspace{0.2cm}
\captionsetup{type=figure}
\includegraphics[width=1.0\linewidth]{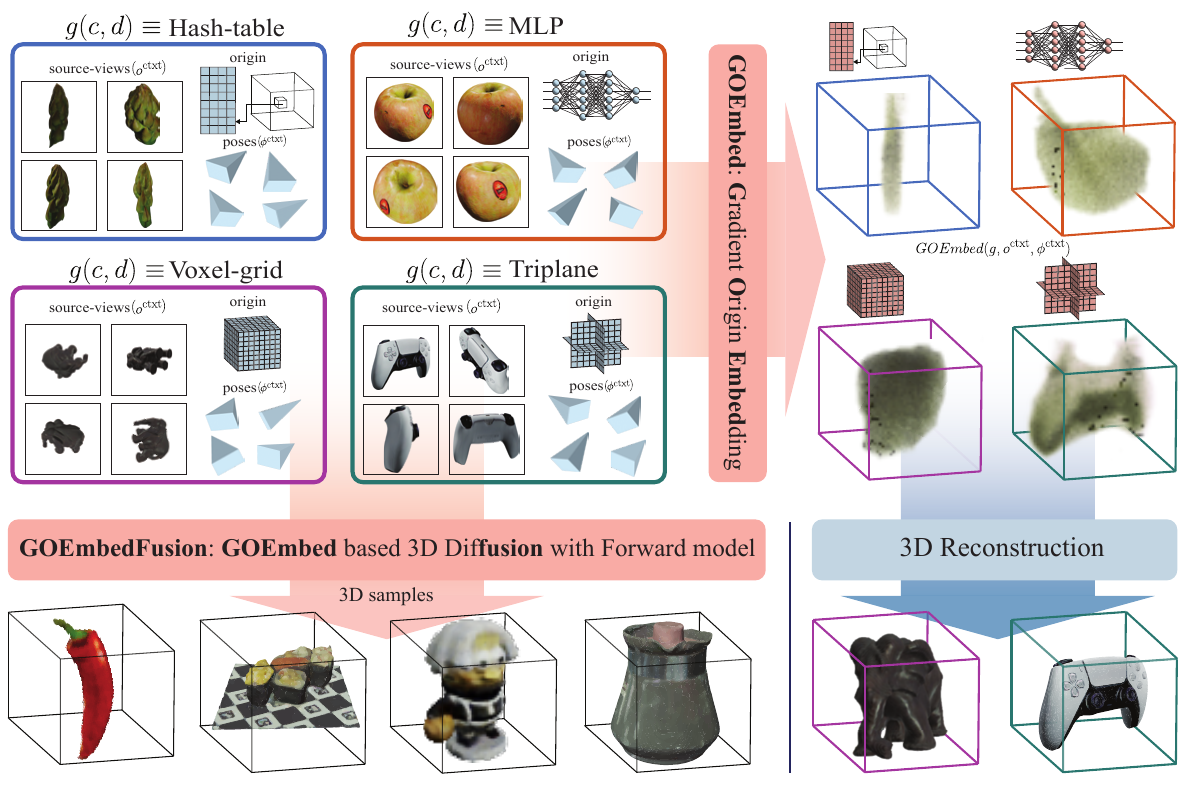}
\captionof{figure}{We propose the \textbf{\name}~(\textbf{G}radient \textbf{O}rigin \textbf{Embed}ding) mechanism that encodes source views ($o^\text{ctxt}$) and camera parameters ($\phi^\text{ctxt}$) into arbitrary 3D Radiance-Field representations $g(c,d)$ (sec.~\ref{sec:method}). We show how these general-purpose {\name}dings can be used in the context of 3D DFMs (Diffusion with Forward Models) (sec.~\ref{sec:3d_gen}) and for sparse-view 3D reconstruction (sec.~\ref{sec:3d_recon}).}%
\label{fig:teaser-fig}
\end{center}

\input{src/0_abstract}

\input{src/1_intro}
\input{src/2_related_work}

\input{src/3_method}
\input{src/4_experiments}

\input{src/5_conclusion}

\input{src/6_acknowledgements}

%
%
\bibliographystyle{bib/splncs04}
\bibliography{bib/holo_diffusion, bib/holo_fusion, bib/vedaldi_general, bib/vedaldi_specific, bib/goen_fusion}

\input{src/X_suppl.tex}

\end{document}

%% file: src/preamble.tex
\usepackage[dvipsnames]{xcolor}


%% file: includes/custom_commands.tex
\usepackage{xspace}
\usepackage{enumerate}

\definecolor{rowblue}{RGB}{220,230,240}
\definecolor{myorchid}{RGB}{150,10,30}
\definecolor{myblue}{RGB}{10,30,250}
\definecolor{mygreen}{RGB}{10,120,10}

\newcommand{\new}[1]{#1}

\definecolor{colorA}{HTML}{aa6600}
\definecolor{colorB}{HTML}{ff9912}
\definecolor{colorC}{HTML}{ffc125}
\definecolor{colorD}{HTML}{7a83e4}
\definecolor{colorE}{HTML}{cdc673}
\definecolor{colorF}{HTML}{cfb53b}
\definecolor{colorG}{HTML}{98a148}
\definecolor{colorH}{HTML}{666666}
\definecolor{colorI}{HTML}{ef5998}
\definecolor{colorJ}{HTML}{9112ff}

\definecolor{colorY}{HTML}{000000}
\definecolor{colorZ}{HTML}{777777}

\colorlet{color$\pi$-GAN}{colorA}
\colorlet{colorEG3D}{colorB}
\colorlet{colorGET3D}{colorC}
\colorlet{colorStable-DreamFusion}{colorD}
\colorlet{colorDreamFusion}{colorD}
\colorlet{colorHoloDiffusion}{colorE}
\colorlet{colorHoloDiffusion$^*$}{colorF}
\colorlet{colorHoloFusion}{colorG}
\colorlet{colorHoloFusion (Ours)}{colorG}
\colorlet{colorOurs}{colorG}
\colorlet{colorOur}{colorG}
\colorlet{colorHoloFusion (MSE)}{colorY}
\colorlet{colorMSE}{colorY}
\colorlet{colorHoloFusion (SDS)}{colorY}
\colorlet{colorSDS}{colorY}
\colorlet{colorHoloFusion (w/o Patch remix)}{colorY}

\definecolor{tabfirst}{rgb}{1, 0.7, 0.7} 
\definecolor{tabsecond}{rgb}{1, 0.85, 0.7} 
\definecolor{tabthird}{rgb}{1, 1, 0.7} 

\newcommand{\gon}{GON}

\newcommand{\ffwddiff}{Forward-Diffusion}

\newcommand{\name}{GOEmbed\xspace}


%% file: src/0_abstract.tex
\begin{abstract}
Encoding information from 2D views of an object into a 3D representation is crucial for generalized 3D feature extraction. \new{Such features can then enable 3D reconstruction, 3D generation, and other applications.}
We propose \name (Gradient Origin Embeddings) that encodes input 2D images into any 3D representation, \textit{without} requiring a pre-trained image feature extractor;
unlike typical prior approaches in which
input 
images are either encoded using 2D features extracted from large pre-trained models, or customized features are designed to handle different 3D representations; or worse, encoders may not yet be available for specialized 3D neural representations such as MLPs and hash-grids.
We extensively evaluate our proposed 
\name under different experimental settings on the OmniObject3D benchmark.
First, we evaluate how well the mechanism compares against prior encoding mechanisms on multiple 3D representations using an illustrative experiment called Plenoptic-Encoding. 
Second, the efficacy of the GOEmbed mechanism is further demonstrated by achieving a new SOTA FID of 22.12 on the OmniObject3D generation task using a combination of \name~and DFM (Diffusion with Forward Models), which we call {\name}Fusion.
Finally, we evaluate how the \name mechanism bolsters sparse-view 3D reconstruction pipelines.

\end{abstract}

%% file: src/1_intro.tex
\section{Introduction}
\label{sec:intro}

The rate of progress in 3D Computer Vision research has risen in the last decade due to increased interest in various AR (Augmented Reality), VR (Virtual Reality) and MR (Mixed Reality) applications \cite{tewari2020state, tewari2022advances, xie2022neural, shi2022deep, gao2022nerf}. Many 3D Computer Vision problems are newly being viewed in the light of Deep-Learning based solutions. The process of encoding the information in 2D images into deep features over the chosen 3D representation can be found in the Deep-Learning solutions to almost all these problems, for instance, consider various solutions to long-standing problems such as MVS (Multi-View Stereo) \cite{zhou2018stereo, kar2017learning, nguyen2021rgbd}, NVS (Novel-View Synthesis) or IBR (Image Based Rendering) \cite{hong2023lrm, wang21ibrnet:, yu21pixelnerf:, chen2021mvsnerf, long2022sparseneus}, and 3D reconstruction \cite{wei2020deepsfm, wei2023deepsfm, resindra2018structure}, as well as to the nascent 3D problems such as 3D synthesis \cite{karnewar2023holodiffusion, tewari2023diffusion, chan2023genvs}, and 3D distillation \cite{karnewar2023holofusion, liu2024one}. Surprisingly, despite being such a critical operation, no systematic standalone study of this 2D to 3D encoding operation exists (to the best of our knowledge).

3D scenes/assets do not have a \textit{de-facto} data representation, and different representations are utilized depending upon the requirements of the applications. For instance, just for representing Radiance Fields of static 3D assets, various neural data representations such as MLPs \cite{mildenhall20nerf:, barron2021mip, sitzmann19scene}, Triplanes \cite{chan2022efficient, chen2022tensorf}, Feature-voxel-grids \cite{lombardi2019neural, liu22neural}, Hash-grids \cite{muller2022instant, takikawa2022variable}, as well as non-neural ones like ReLU-Fields \cite{karnewar2022relu}, Plenoxels \cite{yu2021plenoxels}, DVGo \cite{sun2022direct}, 3DGS \cite{kerbl3Dgaussians} are being utilized in various 3D applications. 
Thus, given this disarray around 3D scene representations, it is a key challenge to come up with a 2D-to-3D encoding method that can: (i)~generalize to arbitrary 3D representations, (ii)~while being able to capture maximum information in the 3D features from the 2D images.

Existing methods of encoding can be grouped into two categories. (i)~In the first category, the methods are similar to cost-volume-construction-like approaches where 2D deep features are extracted from the images, and then these 2D features are un-projected into the 3D space. On top of the deep-feature extraction network, this un-projection operation can require predicted depths using off-the-shelf depth-estimation models \cite{birkl2023midas, bhat2023zoedepth, yang2024depth} for 3D representations such as point-clouds \cite{wiles2020synsin, aliev2020neural, rakhimov2022npbg++}. For 3D representations such as feature-voxel-grids, \textit{per-voxel-costs} in the form of variance of the per-2D-view features implicitly encodes 3D depths but requires large amount of compute-memory \cite{nguyen2021rgbd, zhou2018stereo}. What is further limiting is that the un-projection operation, is non-trivial and challenging to extend to specialized 3D neural representations such as MLPs and Hash-grids, and hence the approaches from this category are mostly limited to only certain 3D representations. (ii) In the second category, the methods entirely circumvent all forms of 3D inductive biases and directly inject the 2D features into the problem specific backbone networks. For example, methods like NeRFformer \cite{reizenstein21common} and LRM \cite{hong2023lrm} use cross-attention to directly inject 2D features into the 3D sparse-view-reconstruction backbone network. Apart from the limitation of requiring the memory-heavy cross-attention operation, given the results of our sparse-view-3D-reconstruction experiments (sec. \ref{sec:3d_recon}), we hypothesize that these learned-feature-injection approaches may not be learning sufficient 3D-priors and are specializing only to the domains of training (albeit on a large-scale).

To overcome the aforementioned limitations, we propose \textbf{\name}: \textbf{G}radient \textbf{O}rigin \textbf{Embed}dings (fig.~\ref{fig:teaser-fig}) to encode the information in 2D images into any 3D representation which exists currently or which will be proposed in the future; \new{as long as a differentiable \textit{render} operation can be defined on it}. Succinctly, the \name~defines the 3D embeddings as the gradient of the mean-squared-error between renders (of the origin 3D representation) and the G.T. 2D views wrt. a predefined origin over the chosen 3D representation (fig. \ref{fig:method} and sec. \ref{sec:method}). In most cases (except for MLPs due to symmetry-breaking), a simple zero-feature initialization is sufficient to define the origin. Apart from the 3D representation agnostic 
applicability, our {\name}dings are light-weight since they do not require memory-heavy large pretrained 2D feature-extraction networks, and they try to maximize the information transfer between 2D and 3D. In summary, our contributions are:
\begin{itemize}
    \item We propose the \textbf{\name} as a representation agnostic encoding mechanism of 2D source views into different 3D representations (sec. \ref{sec:method}, eq. \ref{eq:goembed_enc}).
    \item We propose a novel 3D diffusion pipeline by combining our \name~with DFM (Diffusion with Forward Models) to achieve the state-of-the-art score on the OmniObject3D generation benchmark (sec. \ref{sec:3d_gen}, eq. \ref{eq:goembed} - \ref{eq:denoise_diff_goembed_recon}).
    \item We evaluate the efficacy of \name~for extracting different 3D representations from source images in the illustrative Plenoptic Encoding experiment (sec. \ref{sec:plen_enc}); and also evaluate its utility in the sparse-view 3D reconstruction task (sec. \ref{sec:3d_recon}).
\end{itemize} 

%% file: src/2_related_work.tex
\section{Related Work}
\label{sec:related_work}

Our proposed method is primarily an attempt at a systematic study of 2D-to-3D encoding mechanisms, thus we firstly cover the prior encoding methods in subsec. \ref{sec:related_work:subsec:2d_to_3D_encoding}. Secondly, since our method can encode 2D into arbitrary 3D representations, we cover the works related to neural 3D scene representations in subsec \ref{sec:related_work:subsec:3d_neural_reprs}. And lastly, since we propose a new realisation of the DFM framework, we cover prior 3D generative modeling works in subsec \ref{sec:related_work:subsec:3d_gen}. Please refer to the survey works of Xie et al. \cite{xie2022neural} and Shi et al. \cite{shi2022deep} for more exhaustive coverage on 3D neural representations and 3D generative models, respectively.

\subsection{2D-to-3D Encoding} 
\label{sec:related_work:subsec:2d_to_3D_encoding}
Most of the 3D encoding mechanisms proposed till now have been in the context of a larger problem such as MVS or NVS, and there has not been a standalone principled study of the encoding mechanisms yet. Nevertheless, the early works \cite{gu2020cascade} constructed pixel-disparity based 3D cost-volumes for MVS (Multi-View Stereo) problems. These raw-pixel based cost-volumes were soon superseded by 2D deep image feature based ones \cite{kar2017learning, nguyen2021rgbd, zhou2018stereo, karnewar2023holodiffusion, karnewar2023holofusion} due their memory-compactness and information expressiveness. These approaches typically proceed as: first obtain per-image-per-pixel features using large pretrained image networks such as ResNet \cite{he15deep} or DinoV2 \cite{oquab2023dinov2}; then these features are un-projected into the 3D space using the camera parameters associated with the views. The un-projected features are then accumulated into the feature-voxel grid yielding the cost-volume to be used in various 3D problem contexts. Apart from voxel-grids,  recent works also splat these features on Triplanes \cite{gupta20233dgen, cao2024lightplane}. 
\new{The most similar work to our proposed \name~in terms of idea is the one from Bond-Taylor and Willcocks titled Gradient Origin Networks~(GON)~\cite{bond2020gradient}, even though their proposal has no context of 3D encodings. We take inspiration from GONs, but our proposed {\name}ings are different from them in that: (i)~while the purpose of GON is to obtain a compressed latent space, our \name~is aimed at obtaining a coarse partial estimate of the plenoptic 3D scene given 2D image observations (sec. \ref{sec:method}); and (ii)~our \name~encodings are much more local compared to the latent embeddings obtained using the GON. Please refer to 
the section \ref{supplsec:method:subsec:gon} of 
the supplementary for a discussion of GON.}

\subsection{3D Neural Representations}
\label{sec:related_work:subsec:3d_neural_reprs}
\textit{Neural Fields} movement in the data representations research was spear-headed by the works such as OccupancyNetworks \cite{niemeyer19occupancy} and SRNs \cite{sitzmann19scene} that proposed to use MLPs for representing the 3D shapes as occupancy fields and SDFs (signed-distance fields) respectively. But, most notably, the work NeRF (Neural Radiance Fields) \cite{mildenhall20nerf:} made this idea popular by demonstrating the use of MLPs in the context of an application as complex as 3D Novel-view-synthesis. Concurrently, another important work called SIREN \cite{sitzmann20implicit} showed that sinusoid-activated MLPs could also represent images, videos, MRIs, etc. The research progress exploded after these two works, leading to various works \cite{karnewar2022relu, yu2021plenoxels, sun2022direct, yu2021plenoctrees} proposing voxel-based modifications to the NeRF scene representation; others \cite{garbin21fastnerf:, wang2023f2, muller2022instant} speeding up the training and inference of NeRFs. Numerous works such as Acorn \cite{martel2021acorn}, NSVF \cite{liu2020neural}, Instant-NGP \cite{muller2022instant}, Variable Bitrate Neural Fields \cite{takikawa2022variable}, TensoRF \cite{chen2022tensorf}, and Triplanes \cite{chan2022efficient} mark some of the key hybrid explicit-implicit versions of Neural 3D scene representations. \new{This is but a sampler of the vast distribution of implicit, explicit, and hybrid methods proposed for representing static 3D scenes in the last few years or so. While previous encoding methods only allow 2D image features to be encoded into voxel-grids (or Triplanes via splatting), our proposed method can be applied to encode the features into almost any of these proposed representations. In addition, we believe \name can be applied even to scene representations that will be proposed in the future.}

\subsection{3D Generative Models}
\label{sec:related_work:subsec:3d_gen}
The earliest forms of deep learning based 3D generative models can be traced back to 3D-GAN \cite{wu2016learning}; where prior non-learning based methods of 3D shape reconstruction or synthesis involved template based permutation approaches \cite{kar15category-specific}. Methods evolved to PlatonicGAN \cite{henzler19escaping} and PrGAN \cite{gadelha163d-shape} that generates 3D shapes from an unstructured corpus of uni-category images. More interesting is the approach of HoloGAN \cite{nguyen-phuoc19hologan:} that adds a 3D inductive bias in the form of a 3D feature-grid and a neural renderer to the conventional generator architecture of 2D GAN. Utilizing the differentiable Emission-Absorption renderer component of the NeRF \cite{mildenhall20nerf:}, works like GRAF \cite{schwarz2020graf}, VoxGRAF \cite{Schwarz2022voxgraf}, and PiGAN \cite{chan2021pi} train 3D GANs with better view-consistency using category-specific image datasets. Orthogonal to these approaches, StyleNeRF \cite{gu2021stylenerf} proposed a hybrid approach which utilised a core-3D graf-like base model, followed by a HoloGAN-like neural renderer to upsample the rendered resolution. Surprisingly, due to carefully crafted architecture and strong regularization, StyleNeRF generated samples hold formidable 3D view-consistency. Simultaneously various works \cite{karnewar20223ingan, zhang2023seeing, wu2023sin3dm, wu2022learning, son2023singraf, li2023patch} proposed GAN and Diffusion based 3D generative models on single-3D scene setting. This progress continued steadily till the recent milestone works in 3D generative models such as EG3D \cite{chan2022efficient}, which trains a 3D GAN on newly proposed 3D Triplane neural representation; GET3D \cite{gao2022get3d}, which uses DMTet \cite{shen2021deep} module in the GAN to directly generate 3D textured meshes; DiffRF \cite{muller2022diffrf}, which proposed the first preliminary 3D diffusion model on 3D voxels similar to 3D-GAN \cite{wu2016learning}; works such as RenderDiffusion~\cite{anciukevicius22renderdiffusion:}, HoloDiffusion~\cite{karnewar2023holodiffusion}, DFM \cite{tewari2023diffusion} which train 3D diffusion models only using 2D images; until the most recent DiffTF \cite{cao2023large} that adapted the DiffRF training pipeline to Triplanes and improved it further to achieve the state-of-the art results on the OmniObject3D \cite{wu2023omniobject3d} dataset. \new{Our proposed {\name}Fusion pipeline is quite similar to DFM \cite{tewari2023diffusion}, but differs from it in the first step. While DFM used the voxel-grid specific feature-extractor, {\name}Fusion uses \name for encoding the features from 2D Images into any chosen 3D scene representation, making it more generally applicable.}

%% file: src/3_method.tex
\section{Method}
\label{sec:method}

\begin{figure}
\centering
\includegraphics[width=\linewidth]{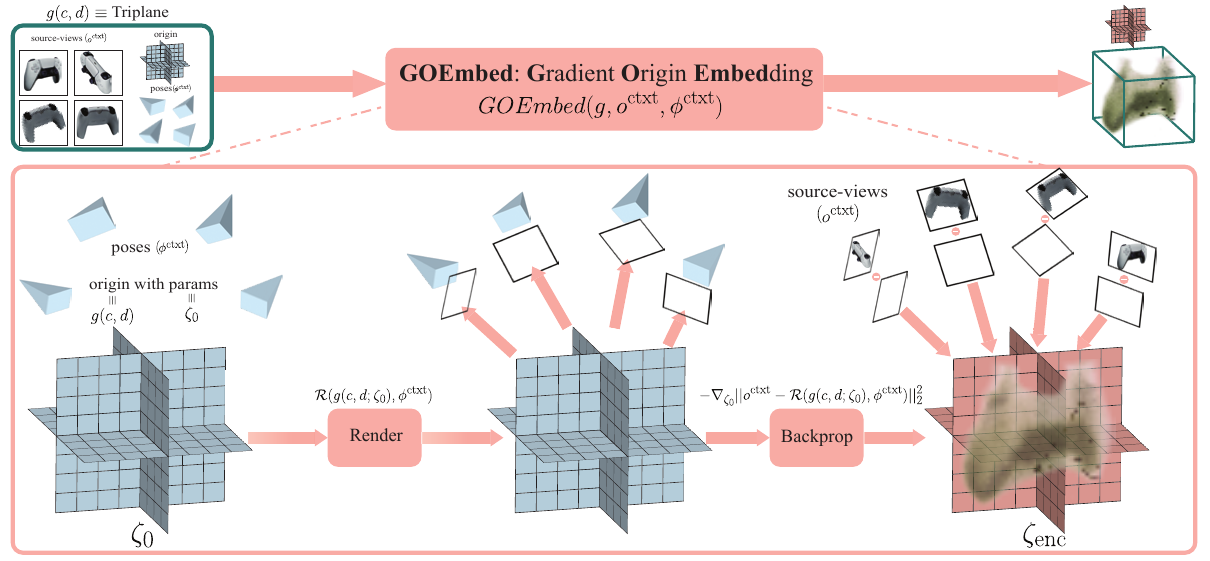}
\caption{
\textbf{\name~illustration}. We demonstrate the mechanism here using the Triplane representation for $g(c, d)$, but note that this can be applied to other representations as well. The \name~mechanism (eq. \ref{eq:goembed_enc}) consists of two steps. First we render the origin $\zeta_0$ from the context-poses $\phi^\text{ctxt}$; then we compute the gradient of the MSE between the renders and the source-views $o^\text{ctxt}$ wrt. the origin $\zeta_0$ which gives us the \name~encoding $\zeta_\text{enc}$.
\label{fig:method}
}
\end{figure}

\subsection{\name: Gradient Origin Embeddings}
Let $g(c, d)$ represent a static 3D scene as a Radiance-Field such that $c=[x, y, z]$  
denotes the 3D coordinates of a point in the Euclidean space, $d=[\theta, \gamma]$\footnote{We deviate from the more common use of $[\theta, \phi]$ for spherical polar coordinates in favour of $[\theta, \gamma]$ to avoid confusion with $\phi^\text{ctxt}$ and $\phi^\text{trgt}$ used to denote context and target camera parameters respectively.} denotes the spherical polar coordinates of an outgoing direction from the point, and the function $g: \mathbb{R}^5 \rightarrow \mathbb{R}^4$ maps each 3D coordinate and a particular outgoing direction to four values ($\sigma, R, G, B$). Here, $\sigma$ represents density and $[R,G,B]$ represents the outgoing radiance.
Let $\mathcal{R}(g, \phi)$ denote the rendering \textit{functional} that converts the Radiance-Field function into an
image of a certain resolution from a certain camera viewpoint, as described by the camera parameters $\phi$. 
We use the differentiable Emission-Absorption Volume Raymarching algorithm for rendering \cite{mildenhall20nerf:,max1995optical}. 
For the encoding mechanism to be unified and general-purpose, it
must have the following three properties:
\begin{enumerate}[(i)]
    \item It should be able to encode one or more views alike.
    \item It should generalize to any parameterization/realization of the function $g$,  
    i.e., the encoding mechanism should be applicable irrespective of whether the Radiance-Field is represented as an MLP \cite{mildenhall20nerf:}, a Hash-grid \cite{muller2022instant}, a Triplane \cite{chan2022efficient, wang2022rodin}, or a Voxel-grid \cite{karnewar2022relu, yu2021plenoxels, sun2022direct}.
    \item It should try to maximize the information transfer from the 2D views to the 3D embedding.
\end{enumerate}
We introduce the Gradient Origin Embeddings, where we define the encoding of the observations (2D views) as the gradient of the mean-squared error between the rendered and ground truth 2D views with respect to a predefined origin (zero vector or features) 3D representation.

Without any loss of generality, assuming that $\zeta$ are the parameters of the $g$ function (i.e., the features/weights of the 3D Radiance-Field) and $\zeta_0$ denotes the origin (zero features/weights), we define the encodings $\zeta_\text{enc}$ as (fig.~\ref{fig:method}):
\begin{align}
    \zeta_\text{enc} &:= \name(g, o^\text{ctxt}, \phi^\text{ctxt}) \nonumber \\
                     &\boxed{:= -\nabla_{\zeta_0}||o^\text{ctxt} - \mathcal{R}(g(c,d;\zeta_0), \phi^\text{ctxt})||^2_2} \label{eq:goembed_enc}
\end{align}
where, $o^\text{ctxt} \text{ and } \phi^\text{ctxt}$ are the G.T.~2D views and their camera parameters, respectively, which are to be encoded into the 3D representation $\zeta_\text{enc}$.
Note that the proposed encoding function \name backpropagates through the differentiable rendering functional $\mathcal{R}$. By construction, such an encoding satisfies the formerly stated properties (i) and (ii). Further, we minimize the loss function
\begin{align}
    \mathcal{L}^\text{\name}(o^\text{ctxt}, o^\text{trgt}) & := ||o^\text{ctxt} - \hat{o}^\text{ctxt}  ||^2_2 + ||o^\text{trgt} - \hat{o}^\text{trgt}  ||^2_2 \nonumber \\
    \text{where, }
    \hat{o} &= \mathcal{R}(g(c, d; \zeta_\text{enc}), \phi),
\end{align}
for maximizing the information content in the encodings $\zeta_\text{enc}$ to satisfy property~(iii). Here, $\phi^\text{ctxt}$ are the context views used for encoding while $\phi^\text{trgt}$ are some target views of the same scene but different from the source ones. Intuitively, we repurpose the backward pass of the rendering functional to encode the information in the source views $o^\text{ctxt}$ into the parameters of the 3D scene representation $\zeta$. \new{Thus, as long as a mathematically differentiable rendering operator is possible on it,} any 3D scene representation can  be encoded using our \name~encoder.

\subsection{Experimental Evaluation Rubric}
\label{sec:experiments}
We evaluate the generality of \name~mechanism along three axes: firstly, to measure  
the information transfer, we run experiments in a Plenoptic Encoding setting (sec.~\ref{sec:plen_enc}); secondly, we train the {\name}Fusion model to learn a 3D generative model using 2D images to evaluate its 3D generative capability (sec.~\ref{sec:3d_gen}); and finally, we also evaluate the \name~mechanism in a sparse-view 3D reconstruction setting (sec.~\ref{sec:3d_recon}).

\paragraph{Dataset.} We perform all our experiments on the recently released OmniObject3D dataset \cite{wu2023omniobject3d} containing $\sim$6K 3D scans of real world objects from daily life. The dataset contains a large-vocabulary of $\sim$200 categories of daily life classes having some overlap with COCO \cite{lin2014microsoft}.
Only for our non-forward diffusion baseline, 
we also use the text-captions recently released by the OmniObject3d authors at their GitHub page \cite{omni-object3d-captions}.  

\paragraph{Metrics.} For the experiments in the Plenoptic-Encoding setting (sec.~\ref{sec:plen_enc}), we use the standard image reconstruction metrics \textbf{PSNR}, \textbf{LPIPS} \cite{zhang2018unreasonable} and \textbf{SSIM} \cite{wang2003multiscale}, while for the quantitative analysis in generative modeling experiments (sec.~\ref{sec:3d_gen}) we use the \textbf{FID} \cite{heusel2017gans} and \textbf{KID} \cite{binkowski2018demystifying} metrics following prior works.
And, lastly for the 3D reconstruction experiments (sec.~\ref{sec:3d_recon}) we again use the standard image reconstruction metrics similar to the Plenoptic-Encoding experiments. We color code all the scores as \colorbox{tabfirst}{first}, \colorbox{tabsecond}{second}, and \colorbox{tabthird}{third}. 

\begin{figure}
\centering
\includegraphics[width=0.71\linewidth]{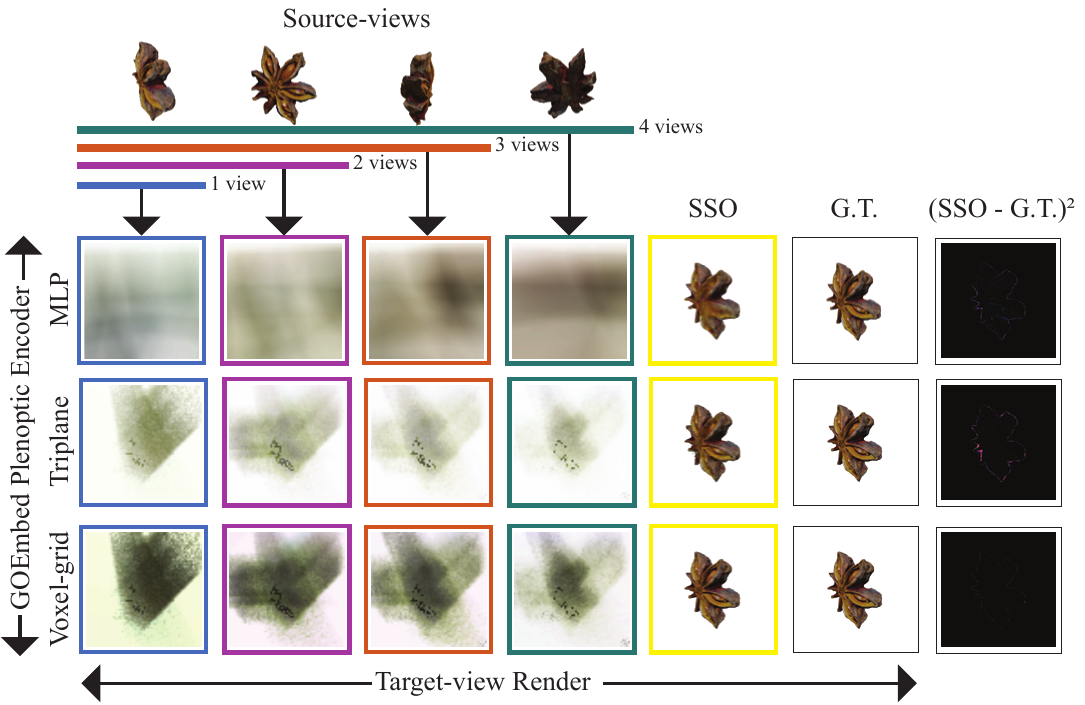}
\caption{
\textbf{Plenoptic Encoding Qualitative Evaluation}. \new{The rows MLP, Triplane and Voxel-grid show the renders of the \name~encoded representations from the target-view respectively, depicting the varying amounts with which \name is capable of encoding the 2D source information into 3D. The colour-coded columns demonstrate the effect of varying the number of source-views (top row) used in the \name~encoding, viz. 1, 2, 3, and 4. The SSO column shows the target render of the single-scene-overfitted representation while the G.T. column shows the mesh-render from the dataset (repeated for clarity). The rightmost column visualizes the pixelwise squared error between the G.T. and the SSO.} 
\label{fig:partial_autoencoding_qualitative}
}
\end{figure}

%% file: src/4_experiments.tex
\section{Plenoptic Encoding}
\label{sec:plen_enc}

\input{src/tables/plen_enc_quant_table}

To evaluate the information transfer from the source views to the encoded 3D representation, we train the standalone \name~component on its own before using it in different contexts. Specifically, given the dataset $D = \lbrace(\mathcal{I}_i^j, \phi_i^j) | i \in [1, N] \text{ and } j \in [1, C]\rbrace$ of $N$ 3D scenes where each scene contains $C$ images and camera parameters, we define the Plenoptic Encoding as a mechanism which encodes $k$ source views and camera parameters, of a certain 3D scene, into the representation $g$ (whose parameters are $\zeta$). The encoded scene representation should be such that the rendered views 
from the same source cameras, and some $l$ different target cameras, should be as close as possible to the G.T. images $\mathcal{I}$, i.e., the PlenopticEncoder $PE: \mathbb{R}^{k \times h \times w \times c} \times \mathbb{R}^{k \times 4 \times 4} \rightarrow \mathbb{R}^{(k + l) \times h \times w \times c}$ should minimize the following mean squared error objective:
\begin{align}
    \mathcal{L}^\text{PE-MSE} &:= \mathbb{E}_{(\mathcal{I}, \phi) \sim D}\| \mathcal{I} - PE(\mathcal{I}, \phi) \|^2_2 \nonumber.  
\end{align}
Although this experimental setup is quite similar to typical MVS/NVS or 3D reconstruction or 3D prior learning setups, in form and essence; we note here that the plenoptic encoder $PE$ 
is neither targeted to do Multi-View Stereo nor 3D reconstruction. To set the correct expectations, we note that the main and the only goal here is to evaluate how much information is transferred from the 2D images to the chosen 3D representation. We evaluate the \name~encodings on three different 3D Radiance-Field representations, $g$, namely Triplanes \cite{chan2022efficient}, Feature-Voxel grids \cite{karnewar2022relu,karnewar2023holodiffusion, yu2021plenoxels, sun2022direct}, and MLPs \cite{mildenhall20nerf:} and compare them to cost-volume like approaches where possible.

All the evaluations in Table~\ref{tab:plen_enc_quant} are done on $100$ randomly chosen \textit{test} scenes from the OmniObject3D dataset  on $1$, $2$, $3$, and $4$, number of source views. Although the loss is optimized on both source and target camera views, the reported scores are for \textit{target-views} only, since we are interested in measuring the ``3D'' encoding ability, as opposed to overfitting the source views into the representation. If we consider only the source-view metrics, there is a possibility of the degenerate solution where the encoder copies all the source views as-it-is disregarding any 3D structure. Also, all the scores are computed against the renders of the G.T. meshes, but since some quality can be lost by the choice of the representation itself, the SSO (Single-Scene-Overfitting) version, where we fit the representation directly to the scene, is also provided for a more grounded comparison.

Table \ref{tab:plen_enc_quant} summarizes the scores obtained in this setup of experiments. We compare against cost-volume construction approach using DinoV2 \cite{oquab2023dinov2}, where we first un-project the per-view image features into a feature-voxel grid and then accumulate the per-view features into the cost-volume similar to StereoMachines \cite{kar2017learning} and many others \cite{nguyen2021rgbd, karnewar2023holodiffusion, chen2021mvsnerf, liu2024one, zhou2018stereo}. We further splat the voxel-features into the mutually orthogonal planes in case of the Triplanes based cost-volume baseline. As apparent from the table, \name~outperforms the DinoV2 cost-volume-approach for Triplanes and comes very close to the Voxel-Grids one. Note that the cost-volume approach is not trivial to come-up with for MLPs, whereas our \name~can handle this representation by design and obtains formidable scores in this setup. Apart from this advantage, \name~only has learnable parameters ($\sim$17K) as part of the renderer functional $\mathcal{R}$ (usually an MLP), while the DinoV2 has excess of $\sim$1B parameters, albeit pretrained. Finally, what is most surprising is that our light-weight \name~approach is actually able to obtain almost 50\% of the SSO scores which is the practically achievable performance for the chosen 3D representations (last row of table \ref{tab:plen_enc_quant}). Figure \ref{fig:partial_autoencoding_qualitative} illustrates this phenomenon visually.

\section{3D Generation}
\label{sec:3d_gen}

\begin{figure}
\centering
\includegraphics[width=0.95\linewidth]{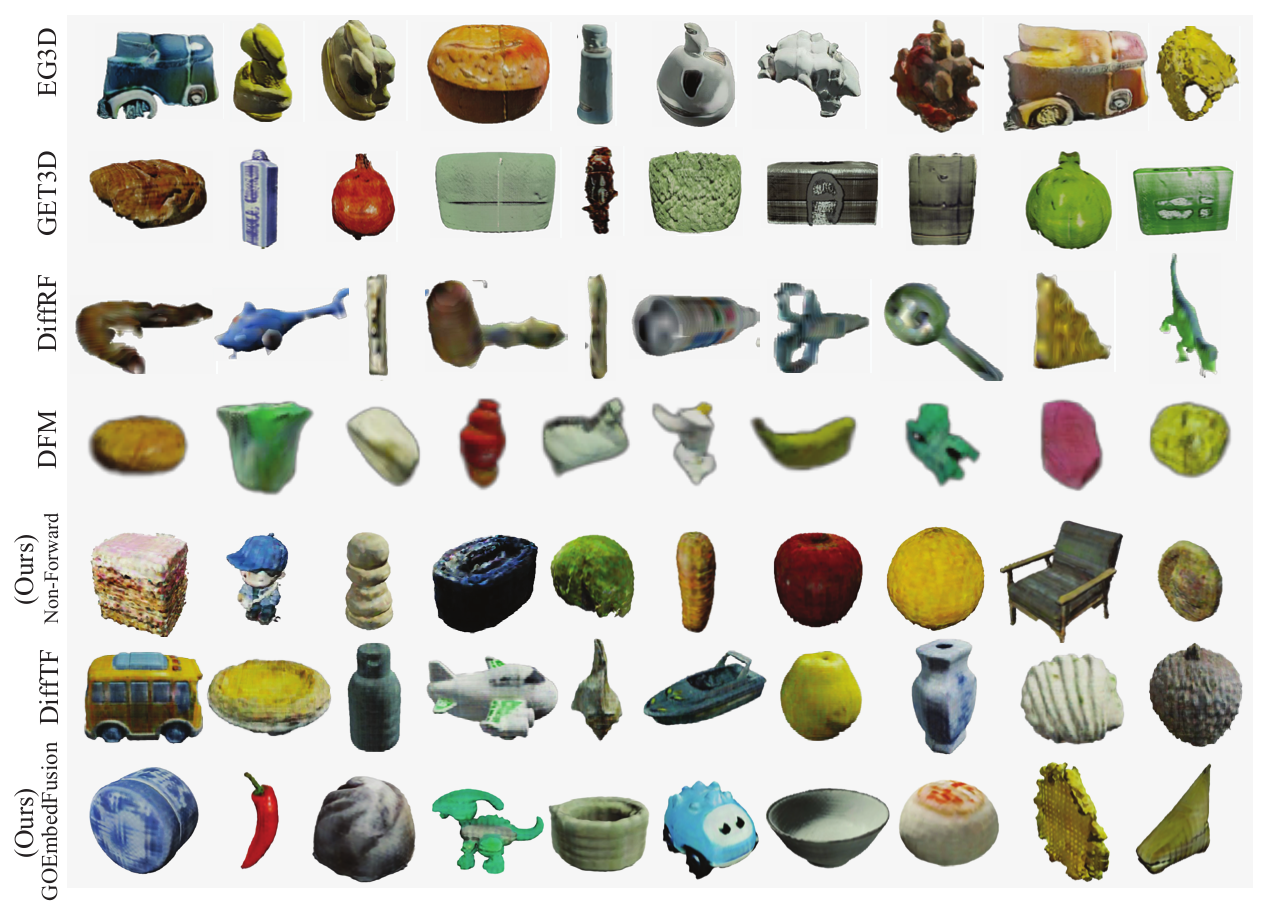}
\caption{
\textbf{3D Generation Qualitative Evaluation}. 3D samples generated by our {\name}Fusion~compared to the prior GAN and Diffusion based baselines. We ran the DFM \cite{tewari2023diffusion} baseline using their code, while used the samples for rest of the baselines from DiffTF \cite{cao2023large}.
}
\label{fig:3d_gen_qualitative}
\end{figure}

Although prior works like RenderDiffusion~\cite{anciukevicius22renderdiffusion:} and HoloDiffusion~\cite{karnewar2023holodiffusion} were already proposed to train 3D diffusion models using only 2D images, recently, the notable work of Tewari et al.~\cite{tewari2023diffusion} proposed a unified mathematical framework for the \textit{stochastic-inverse} setting of generative modeling where we only have access to partial observations of the underlying ground-truth signals, but never the underlying signals themselves; in which 3D inverse graphics is a special case. Please refer to 
section \ref{sec:suppl_prelim:subsec:ffwd_diff} of 
the supplementary for a discussion on the mathematical framework of DFM (Diffusion with Forward Models). Despite this, the proposed vanilla realization of the DFM framework for 3D generative modeling has certain limitations; first, it is only specific to the feature-point-cloud 3D representation and does not generalize to other 3D representations; second, it cannot generate purely unconditional 3D samples since it is a 2D view conditioned 3D diffusion model; and third, it requires a computationally expensive auto-regressive process for sampling the underlying 3D scenes.

We propose a novel realization of the DFM framework called {\name}Fusion where our proposed \name~drives the diffusion with forward model training pipeline; while overcoming all the limitations as mentioned earlier. Our unobserved samples here correspond to the $\zeta$ parameters since we are interested in modeling the generative probability distribution over the 3D Radiance-Fields. A single forward pass through our proposed {\name}Fusion pipeline is defined through the following equations: 

\begin{center}
\boxed{	
	\begin{aligned}
	\zeta_\text{enc} &:= \name(g, o^\text{ctxt}, \phi^\text{ctxt}) \label{eq:goembed} \\
    \hat{\zeta}_0    &:= \mathcal{D}_\theta(\zeta_T, T; \zeta_\text{enc}) \label{eq:goembed_recon} \\
    \hat{\zeta}_t    &\sim q(\hat{\zeta}_t | \hat{\zeta}_{t-1})  = q(\hat{\zeta}_0) \prod_{k=1}^{t} q(\hat{\zeta}_k | \hat{\zeta}_{k-1}) \label{eq:diff_goembed_recon} \\
    \hat{o}^\text{trgt} &:= \mathcal{R}(g(c, d; \mathcal{D}_\theta(\hat{\zeta}_t, t; \zeta_\text{enc})), \phi^\text{trgt})\label{eq:denoise_diff_goembed_recon},	
	\end{aligned}
}
\end{center}
    
where $T$ corresponds to the highest timestep and $\zeta_T \sim \mathcal{N}(0, I)$ is a sample of the standard Gaussian noise distribution. Equation \ref{eq:goembed} first encodes the context observations and their parameters ($o^\text{ctxt}, \phi^\text{ctxt}$) into the \name~encoding $\zeta_\text{enc}$ using eq. \ref{eq:goembed_enc}. Then, eq. \ref{eq:goembed_recon} uses the denoiser network to predict a pseudo-deterministic estimate of the clean 3D scene $\hat{\zeta}_0$ conditioned on the \name~encodings $\zeta_\text{enc}$. 
While the noise can impart some degree of stochasticity to the output, the network can, in theory, completely ignore the noise. We let the network learn what to do based on the training and the data complexity, and hence call this step ``pseudo''-deterministic. This is followed by obtaining a noisy version of the 3D scene $\hat{\zeta}_t$ through standard forward diffusion corruption (eq. \ref{eq:diff_goembed_recon}). We estimate the process of drawing from the distribution $q(\hat{\zeta}_t | \hat{\zeta}_{t-1})$ using the standard DDPM \cite{ho20denoising} closed-form equation $\hat{\zeta}_\text{t} = \sqrt{\alpha_t}\hat{\zeta}_\text{0} + \sqrt{1 - \alpha_t}\epsilon$, where $\epsilon$ is pure Gaussian noise and $\alpha_t$ denotes the schedule of diffusion. And finally, we predict any target observations by rendering the output of the denoiser network $\mathcal{D}_\theta$ for $\hat{\zeta}_t$ conditioned on the \name~encodings $\zeta_\text{enc}$ and the timestep $t$ using eq. \ref{eq:denoise_diff_goembed_recon}.

We note here that the network $\mathcal{D}_\theta$ operates in the \texttt{x\_start} diffusion formulation in contrast to the popular \texttt{epsilon} formulation. We require a few different loss functions to be able to train this pipeline end-to-end such that each component does what it is supposed to:
\begin{align}
    \mathcal{L}^\text{{\name}Fusion} &:= \mathbb{E}_{t,o^\text{trgt}}\|o^\text{trgt} - \hat{o}^\text{trgt}\|^2_2 \nonumber \\
    \mathcal{L}^\text{PSE-DET} &:= \mathbb{E}_{t,o^\text{trgt}}\|o^\text{trgt} - \mathcal{R}(g(c, d; \hat{\zeta}_0)), \phi^\text{trgt})\|^2_2. \nonumber
\end{align}
The final objective 
is simply the sum of the three loss functions as, 
\begin{equation}
    \mathcal{L}^\text{total} := \mathcal{L}^\text{{\name}Fusion} + \mathcal{L}^\text{PSE-DET} + \mathcal{L}^\text{\name}. \nonumber
\end{equation}
The loss function $\mathcal{L}^\text{\name}$ is required to maximize the information content in the \name~encodings. In contrast, the $\mathcal{L}^\text{PSE-DET}$ tries to maximize the reconstruction quality of the pseudo-deterministic output $\hat{\zeta}_0$. The $\mathcal{L}^\text{{\name}Fusion}$ loss function actually trains the {\name}Fusion pipeline. As apparent, the \name~mechanism enables the {\name}Fusion model to use any parameterization of the $g$ function as long as it allows differentiable rendering and can define a zero-origin over its parameters $\zeta$. Please refer to the 
figure \ref{fig:teaser-fig_suppl} in the 
supplementary material for an illustration of the {\name}Fusion model. Empirically, we found  that the two-step bootstrapped-denoising as done in eqn. \ref{eq:goembed_enc} and eqn. \ref{eq:denoise_diff_goembed_recon}, similar to prior works \cite{karnewar2023holodiffusion, karnewar2023holofusion}, is crucial to correctly train this diffusion pipeline in the forward setting. In practice, we train the model using the classifier free guidance training scheme \cite{ho2022classifier}, where we dropout the \name~conditioning randomly with a probability of $0.5$, to allow for both conditional and unconditional sampling. \new{Also, we use the Lighplane \cite{cao2024lightplane} renderer for being able to fit the backward pass 
in GPU memory.} Lastly, samples can be generated by iterative denoising using the trained model $\mathcal{D}_\theta$ either unconditionally or conditionally using the \name~encodings of certain context observations. Here, we can directly sample (denoise) in the space of the unobserved underlying data samples $\zeta$, eliminating the need for the expensive auto-regressive fusion required by DFM. 

\input{src/tables/3d_gen_quant_table}
 
As shown in table~\ref{tab:3d_gen_quant}, our {\name}Fusion sets the new state-of-the-art FID and KID scores on the OmniObject3D dataset in comparison to the prior state-of-the-art DiffTF \cite{cao2023large}, our implementation of a DiffTF like baseline called Ours non-forward, the DFM \cite{tewari2023diffusion} model, the DiffRF \cite{muller2022diffrf} model, and the prior GANs EG3D \cite{chan2022efficient} and GET3D \cite{gao2022get3d}. We note that our improvements are not only limited to the quality of generation, but also to the benefits provided by our {\name}Fusion formulation over the vanilla realisation of the DFM (row 5 table \ref{tab:3d_gen_quant}). While DiffTF \cite{cao2023large} uses many architectural modifications and other tricks specific to Triplanes and the OmniObject3D dataset, we only use the base DiT \cite{peebles2023scalable} architecture with \textit{no} modifications as our backbone denoiser network $\mathcal{D}_\theta$. Also our proposed {\name}Fusion training pipeline is a diffusion with forward model, and hence can be trained only using 2D images, unlike DiffTF.
Our qualitative samples in figure \ref{fig:3d_gen_qualitative} further support these arguments.

\section{Sparse-View 3D Reconstruction}
\label{sec:3d_recon}

The Plenoptic Encoding experiments (sec. \ref{sec:plen_enc}) are in an illustrative setup and provide insights into how well our propsed \name~can transfer information from 2D images into various different 3D representations such as Triplanes, MLPs and Voxel-Grids. In this section, we aim to evaluate the utility of the {\name}dings in a practical application setup. Thus, although the mathematical experimental setup is similar to the Plenoptic Encoding one, here we input the obtained {\name}dings to a backbone sparse-view-3D-reconstruction network. Specifically, we use the DiT based transformer network, which is 12-layers wide as the reconstruction backbone. The end-to-end pipeline is supervised with exactly the same losses as that of the Plenoptic Encoding setup. Intuitively, our \name~reconstruction model replaces the input learned triplane positional encodings with our {\name}dings and removes the cross-attention layers from the LRM (Large Reconstruction Model) architecture.
Additionally, we also evaluate the reconstruction capability of our {\name}Fusion model by running the pipeline only till the ``pseudo''-deterministic output stage (i.e., eqn. \ref{eq:goembed_recon}).

We compare this transformer-based reconstruction setup of ours to the most recent baseline of LRM (Large Reconstruction Model) \cite{hong2023lrm}, which is also based on the transformer architecture. Since LRM's code has not been published, we implement this baseline in our code-base as close to the paper-description as possible, for a fair comparison. Specifically, we train two versions of the LRM, the first one which is 16-layers wide (the base published model), and second smaller version which is 6-layers wide. We introduced the 6-layers version since we found the base-model to be overfitting to the OmniObject3D training subset, which is much smaller in scale than the dataset on which LRM is trained. All the learned network baselines are evaluated on the $\sim$~200 test scenes of the OmniObject3D dataset while the Triplane-SSO (from the Plenoptic Encoding section \ref{sec:plen_enc}) is evaluated on 100 scenes. These 100 SSO scenes form a proper subset of the test-set, so the comparison here is fair. Please check the supplemental for more details of our {\name} reconstruction architecture.

\input{src/tables/3d_recon_quant_table}

As summarized in the table \ref{tab:3d_recon_quant}, both our \name~reconstruction model as well as the diffusion based {\name}Fusion model outperforms the LRM baselines. Also, our reconstruction model gets quite close to the practical performance limit as set by the Triplane-SSO experiment. Thus, given these results, we can assert that our proposed {\name}dings generalise to datasets like OmniObject3D and can strengthen the sparse-view-3D-reconstruction pipelines. Please refer to the 
supplemetary material for more details.

%% file: src/tables/plen_enc_quant_table.tex
\begin{table*}[h!]
    \centering
    \setlength{\tabcolsep}{0.1cm}
    \captionof{table}{\textbf{Plenoptic Encoding Quantitative Evaluation.} PSNR($\uparrow$), LPIPS($\downarrow$) and SSIM($\uparrow$) reported on three different representations of the 3D Radiance-Field $g$, namely, Triplanes, Voxel-Grids and MLPs. All the metrics are evaluated for target views (different from the source views) against the G.T. mesh renders from the dataset.
    The SSO (Single Scene Overfitting) scores denote the case of individually fitting the representations to the 3D scenes.}
    \resizebox{1.0\linewidth}{!}{%
        \begin{tabular}{r ccc ccc ccc}
             \toprule
             {Method} &
             \multicolumn3c{Triplane} &
             \multicolumn3c{Voxel-Grid} &
             \multicolumn3c{MLP} \\
             
             \cmidrule(lr){2-4}
             \cmidrule(lr){5-7}
             \cmidrule(lr){8-10} 
             
             &
             \multicolumn1c{\footnotesize{PSNR} ($\uparrow$)}&
             \multicolumn1c{\footnotesize{LPIPS} ($\downarrow$)}&
             \multicolumn1c{\footnotesize{SSIM} ($\uparrow$)}&
             \multicolumn1c{\footnotesize{PSNR} ($\uparrow$)}&
             \multicolumn1c{\footnotesize{LPIPS} ($\downarrow$)}&
             \multicolumn1c{\footnotesize{SSIM} ($\uparrow$)}&
             \multicolumn1c{\footnotesize{PSNR} ($\uparrow$)}&
             \multicolumn1c{\footnotesize{LPIPS} ($\downarrow$)}&
             \multicolumn1c{\footnotesize{SSIM} ($\uparrow$)} \\
             \toprule
             DinoV2 1 source-view & 14.182 & 1.286  & 0.425  & 14.419  & 1.151 & 0.485 & N/A & N/A & N/A \\
             DinoV2 2 source-views & 14.482  & 1.235 & 0.435  & 15.330 & \cellcolor{tabthird} 0.925  & \cellcolor{tabthird} 0.526 & N/A & N/A & N/A \\
             DinoV2 3 source-views & 14.664  & 1.213 & 0.430  & \cellcolor{tabthird} 15.685 & \cellcolor{tabsecond} 0.864 & \cellcolor{tabsecond} 0.546 & N/A & N/A & N/A \\
             DinoV2 4 source-views &  14.711  &  1.199 &  0.425 & \cellcolor{tabfirst} 15.834  & \cellcolor{tabfirst} 0.857 & \cellcolor{tabfirst} 0.551 & N/A & N/A & N/A \\ 
             \midrule
             (Ours) 1 source-view & \cellcolor{tabsecond} 15.711 & \cellcolor{tabfirst} 1.025  & \cellcolor{tabfirst} 0.477 & 15.514  & 1.008  & 0.479 & 11.208 & \cellcolor{tabfirst} 1.124 & \cellcolor{tabthird} 0.496 \\ 
             (Ours) 2 source-views & 15.513  & \cellcolor{tabsecond} 1.068 & \cellcolor{tabthird} 0.468 & 15.319 & 1.067 & 0.480 & \cellcolor{tabthird} 11.406 & \cellcolor{tabthird} 1.197 & 0.489 \\
             (Ours) 3 source-views & \cellcolor{tabthird} 15.672  & \cellcolor{tabthird} 1.129  & 0.456 & 15.590 & 1.110  & 0.484 & \cellcolor{tabsecond} 11.490 & 1.202 & \cellcolor{tabsecond} 0.500 \\
             (Ours) 4 source-views & \cellcolor{tabfirst} 15.919  & 1.150 & \cellcolor{tabsecond} 0.472 & \cellcolor{tabsecond} 15.755 & 1.145 & 0.486 & \cellcolor{tabfirst} 11.599 & \cellcolor{tabsecond} 1.195 & \cellcolor{tabfirst} 0.505 \\
             \toprule
             \textit{SSO} & 28.165 & 0.087 & 0.941 & 32.061 & 0.067 & 0.9582 & 27.382 & 0.119 & 0.918 \\
             \bottomrule
        \end{tabular}
    }
    \label{tab:plen_enc_quant}
\end{table*}

%% file: src/tables/3d_gen_quant_table.tex
\begin{table}[h!]
    \centering
    \setlength{\tabcolsep}{0.1cm}
    \captionof{table}{\textbf{3D Generation Quantitative Evaluation.} FID($\downarrow$) and KID($\downarrow$) scores on the OmniObject3D dataset comparing our {\name}Fusion ~with GAN baselines EG3D \cite{chan2022efficient}, and GET3D\cite{gao2022get3d}; with the non-forward diffusion baselines DiffRF\cite{muller2022diffrf}, DiffTF\cite{cao2023large}, and Our non-forward diffusion baseline; and, with the DFM (Diffusion with Forward Model) \cite{tewari2023diffusion}.}
    \begin{tabular}{c  cc}
         \toprule
         Method &
         FID ($\downarrow$) &
         KID ($\downarrow$) \\
         \toprule
         EG3D \cite{chan2022efficient}    & 41.56  & 1.0 \\
         GET3D \cite{gao2022get3d}   & \cellcolor{tabthird}49.41  & \cellcolor{tabthird} 1.5 \\
         DiffRF \cite{muller2022diffrf} & 147.59 & 8.8 \\
         DiffTF \cite{cao2023large} & \cellcolor{tabsecond} 25.36  & \cellcolor{tabsecond}  0.8 \\
         \midrule
         DFM \cite{tewari2023diffusion} & 73.51   & 3.8     \\
         Ours non-forward & 119.67  & 8.0  \\
         {\name}Fusion (Ours)  & \cellcolor{tabfirst}22.12 & \cellcolor{tabfirst}0.6      \\
         \bottomrule
    \end{tabular}
    \label{tab:3d_gen_quant}
\end{table}

%% file: src/tables/3d_recon_quant_table.tex
\begin{table}[h!]
    \centering
    \setlength{\tabcolsep}{0.03cm}
    \captionof{table}{\textbf{3D reconstruction Quantitative Evaluation.} PSNR($\uparrow$), LPIPS($\downarrow$) and SSIM($\downarrow$) of our \name~reconstruction model, and {\name}Fusion{'s} ``pseudo''-deterministic 3D reconstruction output compared to LRM baselines. We again include the SSO (Single Scene Overfitting) here for comparison.}
        \begin{tabular}{r  ccc}
             \toprule
             Method &
             
             \multicolumn1c{\footnotesize{PSNR} ($\uparrow$)}&
             \multicolumn1c{\footnotesize{LPIPS} ($\downarrow$)}&
             \multicolumn1c{\footnotesize{SSIM} ($\uparrow$)}\\
             \toprule
             LRM (Our) 6 layer & \cellcolor{tabthird} 23.788 & \cellcolor{tabthird} 0.119 & \cellcolor{tabthird} 0.827 \\
             LRM (Our) 16 layer &  23.247 &  0.121 &  0.813 \\
             \name~recon &\cellcolor{tabfirst}  27.650 &\cellcolor{tabfirst} 0.109 &\cellcolor{tabfirst} 0.900 \\
             {\name}Fusion (PSE-DET) &\cellcolor{tabsecond} 26.447 &\cellcolor{tabsecond} 0.119 &\cellcolor{tabsecond} 0.890 \\
             \toprule
             Triplane \textit{SSO} & 28.165 & 0.087 & 0.941 \\
             \bottomrule
        \end{tabular}
    \label{tab:3d_recon_quant}
\end{table}

%% file: src/5_conclusion.tex
\section{Limitations, Future Scope, and Societal Impact}
Although our framework applies theoretically to any arbitrary 3D Radiance-Field representations, its use in the 3D generative modeling is restricted by the representation's compatibility with existing denoiser network architectures. We believe that Transformers \cite{vaswani17attention} get close to being universally applicable, but it still remains a challenge to adapt certain 3D representations such as Hash-grids \cite{muller2022instant} as input to Transformers. Nevertheless, we believe that our proposed \name~mechanism makes a substantial research stride and will inspire further interesting applications and theoretical insights. Apart from this, upon close qualitative examination, we find that the samples generated using our {\name}Fusion model have some peculiar checkerboard artifacts \new{(refer to 
sec. \ref{suppl_sec:checkerboard_artifacts} of 
the supplementary material)}. Similar to the findings of the recent Denoising-ViT \cite{yang2024denoising}, we hypothesize that these artifacts in our model are also because of the positional encodings in the DiT architecture. Finding the exact reason for these artifacts and getting rid of them constitutes a future direction to be pursued. 

\new{Although our proposed method primarily contributes to extracting 3D features from 2D Images, as shown in our 3D generation experiments,} the 3D features extracted from our method could be applied in the context of the generative modeling of real-captured or synthetic 3D assets. Hence, similar to the case of 2D generative models, our proposed {\name}Fusion model is also prone to the misuse of the synthetically generated media. We note that there is a potential for our method to be used in the creation of fake 3D-consistent videos. 

\section{Conclusion}
\label{sec:conclusion}
We presented the \name~as a mechanism for encoding the information in 2D images into arbitrary 3D scene representations and evaluated its information transfer ability with the Plenoptic Encoding experiments. We show that the encodings can be applied to Triplanes, Voxel-grids and MLPs, but note that exploring these in the context of other popular 3D representations (e.g., meshes, point clouds) forms scope for future work. The {\name}dings find practical usefulness in the context of improving the framework of DFM models, which we demostrate through the 3D generation experiments on the OmniObject3D benchmark; and in the context of sparse-view 3D reconstruction as well.

%% file: src/6_acknowledgements.tex
\section*{Acknowledgements}
\new{Animesh and Niloy were partially funded by the European Union’s Horizon 2020 research and innovation programme under the Marie Skłodowska-Curie grant agreement No.~956585. This research has been partly supported by MetaAI and the UCL AI Centre.}

%% file: src/X_suppl.tex
\clearpage
\setcounter{page}{1}

\title{Supplementary Material for \name: Gradient Origin Embeddings for Representation Agnostic 3D Feature Learning
}

\titlerunning{Supplementary Material for \name}

\author{}
\authorrunning{A.~Karnewar et al.}

\institute{}

\maketitle
\thispagestyle{empty}
\begin{center}
\centering
\captionsetup{type=figure}
\includegraphics[width=1.0\linewidth]{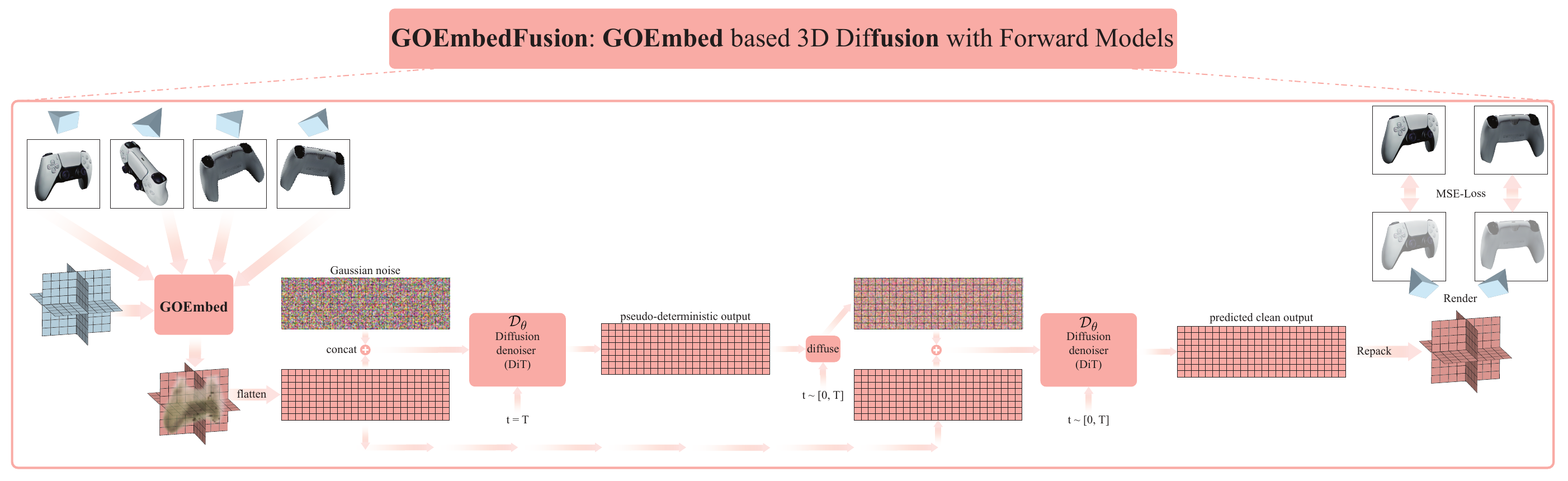}
\captionof{figure}{Illustration diagram for the {\name}Fusion method. Best viewed zoomed in.}%
\label{fig:teaser-fig_suppl}
\end{center}

\section{Preliminaries}
\label{sec:preliminaries}

We start with a summary of the required preliminaries for \textbf{GONs}: Gradient Origin Networks in sec.~\ref{supplsec:method:subsec:gon} and \textbf{Diffusion with Forward} models in sec.~\ref{sec:suppl_prelim:subsec:ffwd_diff}. Please refer to the papers of Bond-Taylor and Willcocks \cite{bond2020gradient}, and Tewari et al. \cite{tewari2023diffusion} for more details about these methods.

\subsection{Preliminaries: \gon{s} (Gradient Origin Networks)}
\label{supplsec:method:subsec:gon}
Given a dataset of observed samples $x\sim p_\text{data}$, where $x \in \mathbb{R}^m$, a Gradient Origin Network \cite{bond2020gradient} (\gon) auto-encodes the input $x$ into reconstructed $\hat{x}$ via a much smaller latent space $z \in \mathbb{R}^k$, where $k \ll m$, without requiring an explicit encoder network. The encoder mapping of the input $x$ to latent $z$, $F_\text{enc}: \mathbb{R}^m \rightarrow \mathbb{R}^k$, is defined as the gradient of the log-likelihood of $x$ under the decoder network $F_\text{dec}: \mathbb{R}^k \rightarrow \mathbb{R}^m$ wrt. a known fixed origin latent $z_0$. In practice, the $z_0$ is always set to zeros, i.e., $z_0 \in \mathbb{R}^k_{[0]}$ and the mean squared error between $F_\text{dec}(z_0)$ and $x$ is used as an estimate of the log-likelihood as follows:
\begin{align}
    z &= F_\text{enc}(x) \nonumber \\
      &= -\nabla_{z_0}\|x - F_\text{dec}(z_0) \|^2_2.
\end{align}
The decoding of $z$ can now simply be obtained as a forward pass of the decoder network $F_\text{dec}$:
\begin{align}
    \hat{x} &= F_\text{dec}(z) \nonumber \\
            &= F_\text{dec}(-\nabla_{z_0}\|x - F_\text{dec}(z_0) \|^2_2).
\end{align}
Using this encoder-less auto-encoding mechanism, the decoder network can be trained using the standard log-likelihood maximization objective (again estimated via mean squared error):
\begin{align}
    L^\text{MSE}(x, \hat{x}) &= \|x - \hat{x}\|^2_2  \nonumber \\
                             &= \|x - F_\text{dec}(-\nabla_{z_0}\|x - F_\text{dec}(z_0) \|^2_2)\|^2_2. \label{eq:gon_loss}
\end{align}

In the \gon~paper \cite{bond2020gradient}, Bond-Taylor and Willcocks also proposed a variational version of the \gon~autoencoder to enable generative sampling of the latent space, a coordinate-based implicit version of the \gon~as a secondary application, and a multi-step generalisation of the Gradient Origin Networks mechanism. But we only describe the auto-encoder version here, because it is the most relevant one to our proposed \name~encodings (sec \ref{sec:method} of main paper).

\subsection{Preliminaries: \textbf{Diffusion with Forward} models}
\label{sec:suppl_prelim:subsec:ffwd_diff}
\textit{Diffusion models} 
define the generative process through two markov chains, namely forward (diffusion) and reverse (denoising), over the observed 
data distribution $p_\text{data}$ 
\footnote{please note that the forward diffusion process here is not to be confused with the forward setting of \textbf{Diffusion with Forward} models.}. The most popular variant DDPM \cite{ho20denoising} defines the forward distribution $q(x_{0:T}) = q(x_0, x_1, ..., x_t, ..., x_T)$ as the markov chain of $T$ Gaussian transitions:
\begin{align}
    q(x_{0:T}) &= q(x_0)\prod^T_1 q(x_t | x_{t-1}) \nonumber \\
    \text{where, } \nonumber \\
    q(x_t | x_{t-1}) &= \mathcal{N}(\sqrt{\alpha_t}x_{t-1}, (1-\alpha_t)I), \label{eq:diffuse}
\end{align}
where the sequence $(\alpha_{1:T})$ instantiates a monotonically decreasing noise schedule such that
the marginal over the last variable approximately equals a standard Gaussian distribution, i.e. $q(x_T | x_{T-1}) \approx q(x_T) \approx \mathcal{N}(0, I)$. 
Similar to the forward process, the reverse process is also defined as another markov chain of $T$ Gaussian transitions: 
\begin{align}
    p(x_{T:0}) &= p(x_T) \prod^{T}_1 p_\theta(x_{t-1} | x_t) \nonumber \\
    \text{where, } \nonumber \\
    p(x_0) &= q(x_0) = p_\text{data}(x) \nonumber \\ 
    p(x_T) &= q(x_T) = \mathcal{N}(0, I) \nonumber \\
    p_\theta(x_{t-1}|x_t) &= \mathcal{N}(\sqrt{\alpha_{t-1}}\mathcal{D}_\theta(x_t, t), (1 - \alpha_{t-1})I) \label{eq:diff_denoise},
\end{align}
where the mean of the Gaussian is given by a learned timestep-conditioned denoiser network $\mathcal{D}_\theta: \mathcal{R}^m \rightarrow \mathcal{R}^m$. 
The network $\mathcal{D}_\theta$ can be trained by minimizing the expectation of the KL divergence between \cite{luo2022understanding} the forward and the reverse transition distributions of a given particular latent $x_t$ over the timesteps $t$ as  
\begin{equation}
    \mathcal{L}_\text{diffusion} = \mathbb{E}_t[D_{KL}(q(x_t|x_{t-1}) || p_\theta(x_t | x_{t+1}))], \label{eq:diffusion_kl}
\end{equation}
and synthetic samples can be drawn from the trained model by iterating $T$-times over eq. \ref{eq:diff_denoise} starting from samples $x_T$ drawn from standard Gaussian distribution $\mathcal{N}(0, I)$. In practice, assuming that the network $D_\theta(x_t, t)$ outputs an estimate of the clean data sample $\hat{x}_0 \approx x_0$ instead of the mean of the Gaussian transition over the previous latent $x_{t-1}$, the KL divergence objective simplifies to 
the mean squared error between $\hat{x}_0$ and $x_0$ \cite{luo2022understanding} 

\begin{equation}
    \mathcal{L}_\text{simple}(x_0, \hat{x}_0) = \mathbb{E}_t[||x_0 - \mathcal{D}_\theta(x_t, t)||^2_2].
\end{equation}

\begin{figure}[t]
\centering
\includegraphics[width=0.95\linewidth]{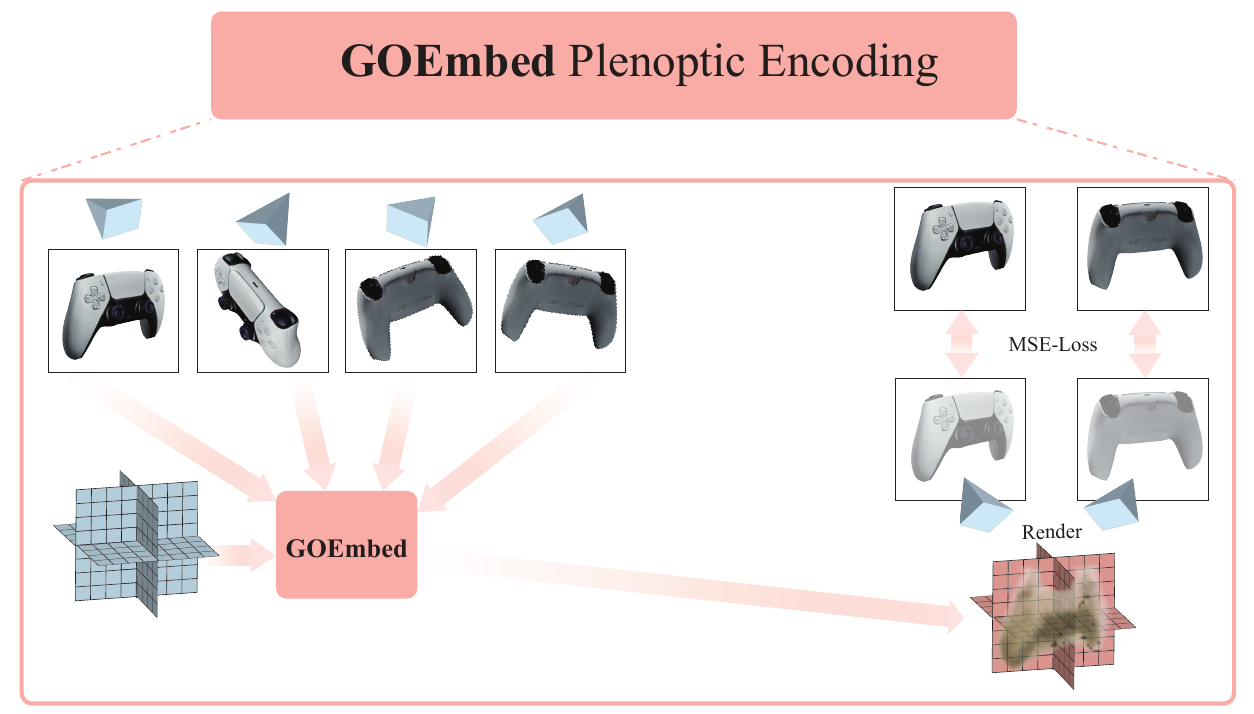}
\caption{
Illustration diagram for the GOEmbed Plenoptic Encoding experimental setup.
}
\label{fig:suppl_goembed_plen_enc}
\end{figure}

\textit{DFM} models \cite{tewari2023diffusion} consider the class of \textit{stochastic-inverse} problems, like inverse 3D graphics, which pose a unique challenge where we do not have direct access to the samples $x \sim p_\text{data}$, but only have access to partial observations of $x$ obtained through a differentiable forward function $o = \texttt{forward}(x, \phi)$, where $\phi$ are the parameters for obtaining the observation. The mathematical framework of DFM 

learns the conditional distribution $p(x | o, \phi)$ over the unobserved data $x$ given the partial observations $(o, \phi)$, by modeling the \textit{pushforward} distribution
\begin{align}
    p_\theta(o^\text{trgt} | o^\text{ctxt}, \phi^\text{ctxt}, \phi^\text{trgt}).
\end{align}
The observations $(o^\text{ctxt}, \phi^\text{ctxt})$ form the context while $(o^\text{trgt}, \phi^\text{trgt})$ are the target observations.
\begin{align}
    &p_\theta(o^{\text{trgt}}_{0:T} | o^\text{ctxt}, \phi^\text{ctxt}, \phi^\text{trgt}) = \nonumber \\ 
    &p(o^{\text{trgt}}_T) \prod^T_1 p_\theta(o^\text{trgt}_{t-1} | o^\text{trgt}_t, o^\text{ctxt}, \phi^\text{ctxt}, \phi^\text{trgt})
\end{align}
In order to 
implicitly model the conditional distribution over the data $x$ from the pushforward, each of the learned reverse transition $p_\theta(o^\text{trgt}_{t-1} | o^\text{trgt}_t, o^\text{ctxt}, \phi^\text{ctxt}, \phi^\text{trgt})$ is defined using the following three steps:
\begin{align}
    x_{t} &= \texttt{denoise}(o^\text{ctxt}, o^\text{trgt}_t, t, \phi^\text{ctxt}, \phi^\text{trgt}) \label{eq:denoise_fwd} \\
    \hat{o}^\text{trgt}_{t} &= \texttt{forward}(x_{t}, \phi^\text{trgt}) \label{eq:fwd} \\
    o^\text{trgt}_{t-1} &\sim \mathcal{N}(\sqrt{\alpha_{t-1}} \hat{o}^\text{trgt}_{t}, (1 - \alpha_{t-1})I). \label{eq:denoise_dist}
\end{align}
Equations \ref{eq:denoise_fwd} and \ref{eq:fwd} define the same functionality as that of the $\mathcal{D}_\theta$ denoiser network, but also integrate the differentiable $\texttt{forward}$ function in the process. 
The noisy versions of the observed target views $o^\text{trgt}_t$ can be obtained by the straightforward diffusion process as defined in eq. \ref{eq:diffuse} without any changes. The \ffwddiff~model can be trained using a conditional version of the KL divergence of eq. \ref{eq:diffusion_kl} as follows:
\begin{align}
    &\mathcal{L}_\text{forward-diffusion} \nonumber \\
    &= \mathbb{E}_t[D_{KL}(q(o^\text{trgt}_t|o^\text{trgt}_{t-1}) || p_\theta(o^\text{trgt}_t | o^\text{trgt}_{t+1}, o^\text{ctxt}, \phi^\text{ctxt}), \phi^\text{trgt})]. \nonumber
\end{align}

And lastly, synthetic samples of the unobserved underlying variable $x$ can be generated in a auto-regressive manner. First we draw a sample $o^\text{trgt}$ starting from a given set of $(o^\text{ctxt}, \phi^\text{ctxt})$ context observations by iterating over equations \ref{eq:denoise_fwd}, \ref{eq:fwd}, and \ref{eq:denoise_dist} $T$ times. A partial estimate of the unobserved underlying variable $x^0_0$ is obtained from the last denoising step.
The complete sample $x$ can be generated by merging $n$ different partial estimates $x^{0:n-1}_0$ by repeating the former process $n$-times and accumulating the drawn $o^\text{trgt}$s together as a new context each time.

\section{{\name}Fusion implementation details}
Fig. \ref{fig:teaser-fig} illustrates the pipeline of the proposed {\name}Fusion method. Although eqs. \ref{eq:goembed}-\ref{eq:denoise_diff_goembed_recon} of the main paper describe the model mathematically, more algorithmic details are provided as follows:
We do all the experiments with the Triplane 3D representation. For the {\name}Fusion model, we use 4 source views and 2 target views for training. We train on the OmniObject3D dataset with a batch size of 16 for $1M$ iterations.
Of all the data and metadata provided in the dataset, we make use of the RGB, depth and normal-map renders of the textured-meshes available as part of the scans and the 3D camera parameters of the rendered views. All the rendered maps in the dataset are natively rendered at the resolution of \texttt{[800 x 800]} in Blender \cite{brito2007blender}. Each of the 3D scan contains $\sim$100 renders from random viewpoints on the upper hemisphere of the 3D centered objects. We use \textit{only} the RGB maps for training the model, while use the depth-maps and normal-maps for computing metrics.

For the diffusion details, we base our code on the guided-diffusion code \cite{guided-diffusion}, and use the default values of $0.0001$ and $0.02$ for the \texttt{beta-start} and \texttt{beta-end} respectively. For the remaining, we use $T=1000$ timesteps, \texttt{cosine} noise schedule and the \texttt{x-start} as the model output. Lastly, we use a single learning rate \texttt{5e-5} for the entire span of the training.

\section{Discussion on 3D reconstruction}

\begin{figure}[t]
\centering
\includegraphics[width=0.95\linewidth]{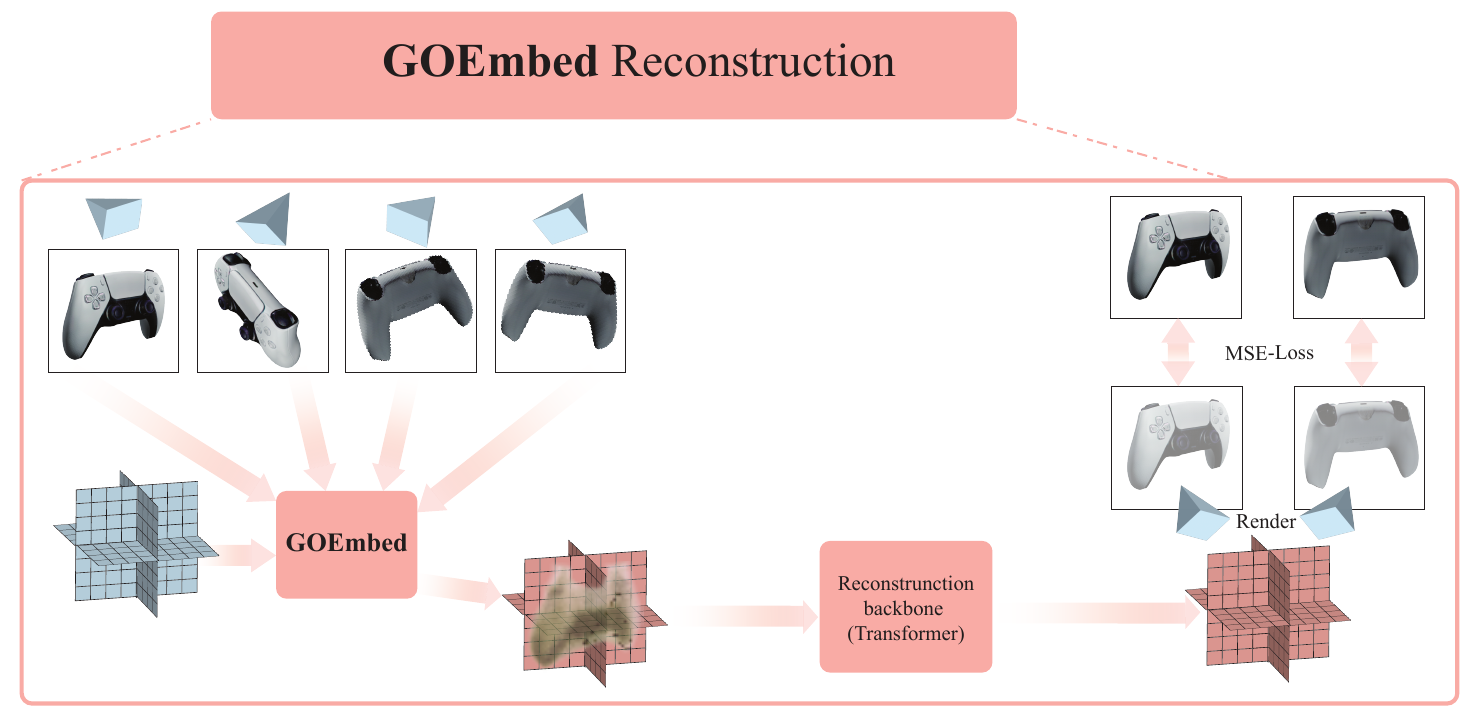}
\caption{
Illustration diagram for the GOEmbed 3D reconstruction experimental setup.
}
\label{fig:suppl_goembed_recon}
\end{figure}

\label{suppl_sec:3d_recon}
Fig. \ref{fig:suppl_goembed_plen_enc} details the setup of the Plenoptic Encoding experiments, while Fig \ref{fig:suppl_goembed_recon} depicts the setup for the sparse-view 3D reconstruction. As we can see, the only difference between these two setups is The presence of a deep learning reconstruction backbone between the two. 
For this baseline, we train a regression based network which reconstructs the 3D scene using only the \name~encoded 3D representation as input. Note that this baseline doesn't apply noise anywhere, and hence gives purely deterministic output. 
As can be seen from the scores of table \ref{tab:3d_recon_quant} of the main paper, the  reconstruction only (regression) model achieves 27.650 dB, while the {\name}Fusion gets as close as 26.447 dB PSNR by itself. Lastly, the Transformer architecture used by us is based on the \texttt{DiT-XL/2} model of Peebles et al. \cite{peebles2023scalable}.

\section{Checkerboard artifacts in generated 3D samples}

\begin{figure}[t]
\centering
\includegraphics[width=0.95\linewidth]{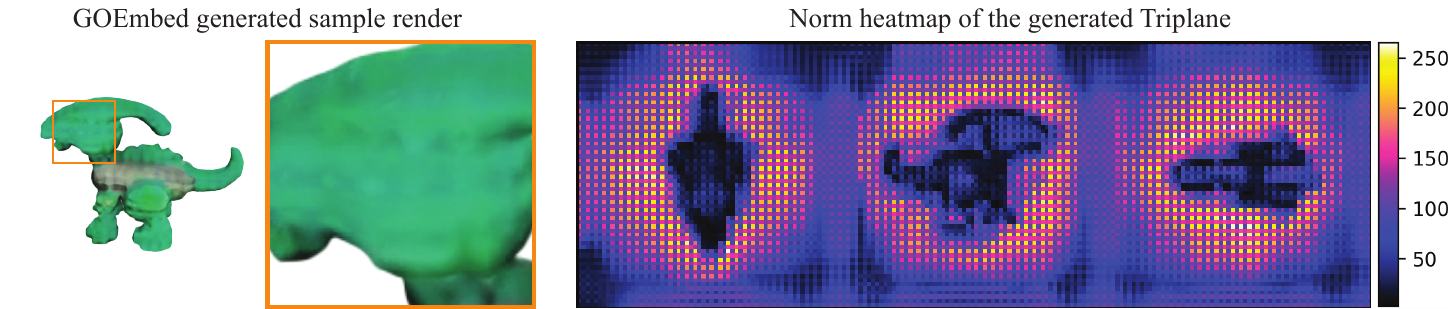}
\caption{Presence of checkerboard artifacts in the {\name}Fusion samples.}
\label{fig:suppl_goembed_checkerboard_artifacts}
\end{figure}

\label{suppl_sec:checkerboard_artifacts}
Fig. \ref{fig:suppl_goembed_checkerboard_artifacts} depicts the presence of checkerboard artifacts in the renders of the generated samples by our {\name}Fusion model. The norm-heatmap of the flattened Triplane features sheds deeper light on the nature of these artifacts. These checkerboard patterns are highly reminiscent of the checkerboard patterns displayed by the recent work of Denoising-VIT \cite{yang2024denoising}, which we believe can be rectified using better positional encodings in the DIT network.